\documentclass{article}

\usepackage[nonatbib,final]{ewrl_2025}  

\usepackage[T1]{fontenc} 
\usepackage[hidelinks]{hyperref}  
\usepackage[utf8]{inputenc}  
\usepackage{algorithmicx}  
\usepackage{algorithm}  
\usepackage{algpseudocode}  
\usepackage{amsfonts}  
\usepackage{amssymb}  
\usepackage{booktabs}  
\usepackage{graphicx}  
\usepackage{hyperref}  
\usepackage{mathtools}  
\usepackage{microtype}  
\usepackage{nicefrac}  
\usepackage{subcaption}  
\usepackage{url}  
\usepackage{xcolor}  

\usepackage[acronym]{glossaries}
\usepackage[style=numeric,sorting=none,backend=biber]{biblatex}
\makeglossaries
\newacronym{ai}{AI}{artificial intelligence}
\newacronym{arma}{ARMA}{auto-regressive moving average}
\newacronym{cpu}{CPU}{central processing unit}
\newacronym{dl}{DL}{deep learning}
\newacronym{fc}{FC}{fully connected}
\newacronym{floris}{FLORIS}{FLOw Redirection and Induction in Steady State}
\newacronym{fnn}{FNN}{feed forward neural network}
\newacronym{gae}{GAE}{generalized advantage estimator}
\newacronym{gat}{GAT}{graph attention network}
\newacronym{gnn}{GNN}{graph neural network}
\newacronym{gpu}{GPU}{graphics processing unit}
\newacronym{gs}{GS}{Gauss-Seidel}
\newacronym{iea}{IEA}{International Energy Agency}
\newacronym{les}{LES}{large-eddy simulations}
\newacronym{lut}{LUT}{lookup table}
\newacronym{mdp}{MDP}{Markov decision process}
\newacronym{mhsa}{MHSA}{multi-head self-attention}
\newacronym{ml}{ML}{machine learning}
\newacronym{mpc}{MPC}{model predictive control}
\newacronym{mw}{MW}{megawatts}
\newacronym{nrel}{NREL}{National Renewable Energy Laboratory}
\newacronym{ppo}{PPO}{proximal policy optimization}
\newacronym{rl}{RL}{reinforcement learning}
\newacronym{sgd}{SGD}{stochastic gradient descent}
\newacronym{td}{TD}{temporal difference}
\newacronym{wffc}{WFFC}{wind farm flow control}

\title{How to craft a deep reinforcement learning policy for wind farm flow control}

\author{%
    Elie Kadoche\textsuperscript{1,2} \quad Pascal Bianchi\textsuperscript{1} \quad Florence Carton\textsuperscript{2} \quad Philippe Ciblat\textsuperscript{1} \quad Damien Ernst\textsuperscript{1,3}\\
    \( ^{1} \)Polytechnic Institute of Paris, 19 Place Marguerite Perey, 91120 Palaiseau, France\\
    \( ^{2} \)TotalEnergies OneTech, 2 Place Jean Millier, 92400 Courbevoie, France\\
    \( ^{3} \)Montefiore Institute, University of Liège, 4000 Liège, Belgium\\
    \texttt{elie.kadoche@ip-paris.fr}\\
}

\begin{document}

\maketitle


\begin{abstract}
	Within wind farms, wake effects between turbines can significantly reduce overall energy production.
	Wind farm flow control encompasses methods designed to mitigate these effects through coordinated turbine control.
	Wake steering, for example, consists in intentionally misaligning certain turbines with the wind to optimize airflow and increase power output.
	However, designing a robust wake steering controller remains challenging, and existing machine learning approaches are limited to quasi-static wind conditions or small wind farms.
	This work presents a new deep reinforcement learning methodology to develop a wake steering policy that overcomes these limitations.
	Our approach introduces a novel architecture that combines graph attention networks and multi-head self-attention blocks, alongside a novel reward function and training strategy.
	The resulting model computes the yaw angles of each turbine, optimizing energy production in time-varying wind conditions.
	An empirical study conducted on steady-state, low-fidelity simulation, shows that our model requires approximately 10 times fewer training steps than a fully connected neural network and achieves more robust performance compared to a strong optimization baseline, increasing energy production by up to 14 \%.
	To the best of our knowledge, this is the first deep reinforcement learning-based wake steering controller to generalize effectively across any time-varying wind conditions in a low-fidelity, steady-state numerical simulation setting.
\end{abstract}


\section{Introduction}\label{sec:introduction}

\subsection{Wind farm flow control}\label{ssec:wind_farm_flow_control}

A wind turbine converts wind energy into electricity.
As wind passes through the blades, its speed decreases, and turbulence increases for a certain distance, creating wake effects.
Within wind farms, wake effects of upstream turbines reduce the power output of downstream turbines (because of lower wind speed) and increase their structural fatigue (because of increased turbulence).
Consequently, wake interactions can significantly decrease overall wind farm energy production.
Wake losses quantify this impact as the percentage of power loss due to wake-induced interference.

Greedy control aims to maximize the power output of each turbine individually by keeping all turbines aligned with the wind direction, without considering wake interactions.
To mitigate the negative impact of wake effects, \gls{wffc}, i.e., coordinated turbine control, can be implemented.
One method, known as wake steering, consists in misaligning upstream turbines in relation to the incoming wind in order to move their wakes away from the downstream turbines.
This is accomplished through yaw control, i.e., active rotation of a turbine’s nacelle around its vertical axis.

\begin{figure}[ht]
	\begin{center}
		\includegraphics[width=0.7\textwidth]{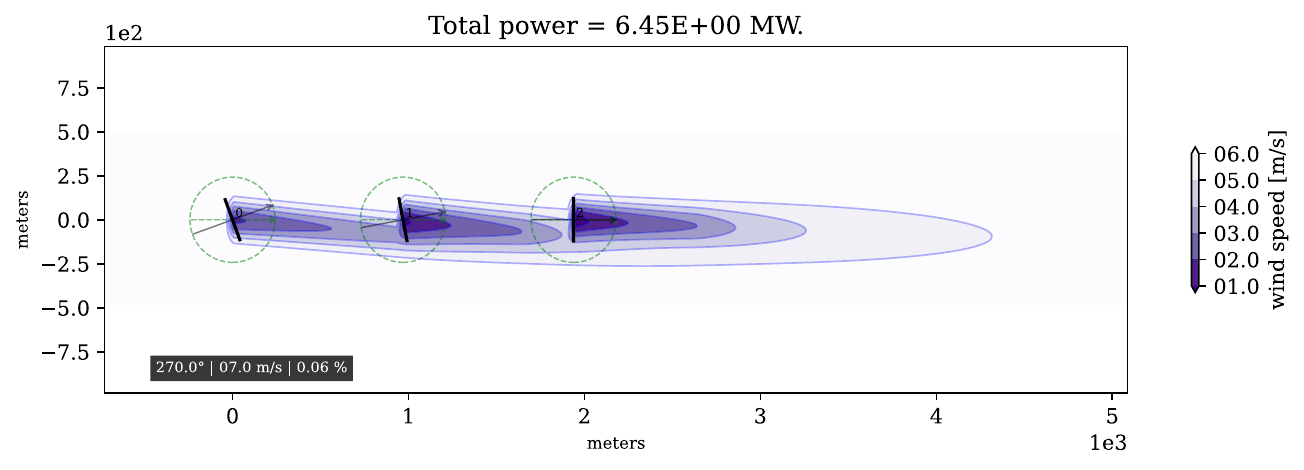}
	\end{center}
	\caption{Example of \gls{wffc} on a three turbines wind farm viewed from above.
		The first two turbines are slightly misaligned with the wind to steer their wake away from the third turbine, maximizing energy production.}\label{fig:wake_effects}
\end{figure}

Figure~\ref{fig:wake_effects} displays a simulation of wake steering applied to a three turbines wind farm with the wind coming from the west.
The darker areas behind each machine correspond to the wake effects.
The first two (upstream) machines are slightly misaligned with the wind direction to redirect their wake away from the third (downstream) turbine.
In this example, the overall farm energy production is improved by 14 \% compared to a standard solution where the three turbines would be aligned with the wind.
However, as wind farms grow larger, the number of wake interactions increases, leading to a high-dimensional control problem with complex spatial dependencies.
In parallel, time-varying and uncertain wind conditions pose additional challenges, as yaw actuators are subject to rotational constraints that limit how quickly and frequently turbines can reorient to shifting flow directions.
Together, these factors make the implementation of \gls{wffc} solutions increasingly challenging.

\subsection{Related works}\label{ssec:related_works}

Wake steering is traditionally implemented using \glspl{lut}, where precomputed yaw settings are applied based on discrete wind conditions~\parencite{field_test_of_wake_steering_at_an_offshore_wind_farm}.
These methods fail to adapt to real-time wind dynamics, leading to suboptimal power production.
Model-based approaches like \gls{mpc}~\parencite{maximum_power_extraction_for_wind_turbines_through_a_novel_yaw_control_solution_using_predicted_wind_directions} offer some adaptability but heavily depend on the accuracy of their underlying wind model and require significant computational resources to solve optimization problems in real time.
To overcome these limitations, data-driven learning-based approaches provide more adaptive and robust control strategies, capable of continuously adjusting yaw settings in response to dynamic and uncertain wind conditions.

In particular, model-free \gls{rl} has emerged as a powerful alternative to traditional wake steering methods~\parencite{data_driven_wind_farm_flow_control_and_challenges_towards_field_implementation_a_review,model_free_closed_loop_wind_farm_control_using_reinforcement_learning_with_recursive_least_squares,actor_critic_agents_for_wind_farm_control}.
Model-free \gls{rl} algorithms are inherently more robust to wind farm model uncertainties and learn solely from experience, enabling the discovery of new and unexpected control strategies, which is especially valuable for large-scale wind farms.
Additionally, \gls{rl} seamlessly integrates multi-objective optimization, making it easy to incorporate fatigue reduction alongside power maximization.

Despite these promising advantages, existing \gls{rl}-based wake steering methods face significant limitations in terms of scalability, sample efficiency, wind dynamics consideration, and wind conditions generalization.
While some studies address one or more of these challenges, none comprehensively tackle all of them together.
Most existing research on \gls{rl} for yaw control assumes quasi-constant wind directions and small wind farms~\parencite{intelligent_wind_farm_control_via_deep_reinforcement_learning_and_high_fidelity_simulations, a_distributed_reinforcement_learning_yaw_control_approach_for_wind_farm_energy_capture_maximization}.
While some studies consider time-varying wind directions, they remain constrained to a limited range~\parencite{deep_reinforcement_learning_for_active_wake_control, marlyc_multi_agent_reinforcement_learning_yaw_control}, leaving uncertainty about their ability to generalize across diverse wind conditions, which is a critical requirement for real-world deployment.
In~\textcite{deep_reinforcement_learning_based_adaptive_yaw_control_for_wind_farms_in_fluctuating_winds}, wind conditions are segmented into discrete intervals, each controlled by a separate \gls{rl} policy.
In this work, we aim to develop a single policy generalizing across all wind conditions.

To enhance learning speed and efficiency, some studies investigate the use of multi-agent \gls{rl} for \gls{wffc}~\parencite{marlyc_multi_agent_reinforcement_learning_yaw_control,falcon_farm_level_control_for_wind_turbines_using_multi_agent_deep_reinforcement_learning}.
However, these approaches often rely on \glspl{fnn}, which struggle to fully capture the complexity of \gls{wffc}.
\Textcite{learning_to_optimise_wind_farms_with_graph_transformers} have recently introduced promising results with the use of graph transformers to build \gls{dl}-based surrogate models for wind power predictions.
In this work, we go further by leveraging \glspl{gat} and self-attention mechanisms within a single-agent \gls{rl} policy, demonstrating significant improvements in sample efficiency, learning performance and generalization.

\subsection{Contributions}\label{ssec:contributions}

In this work, we develop a single-agent deep \gls{rl} policy for wind farm wake steering that generalizes robustly across time-varying and noisy wind conditions.
Our contributions are threefold.
(1) We introduce a novel \gls{rl} architecture combining \glspl{gat} and \gls{mhsa} which improves by about a factor 10 the sample efficiency of a traditional \gls{fnn} and achieves superior performance compared to both a \gls{fnn} and a traditional \gls{gat}.
(2) We propose a new training methodology and reward design that enable the policy to generalize across the full 360\textdegree{} range of wind directions under unsteady and noisy conditions.
(3) And we employ a \gls{ppo} actor-critic framework with a von Mises policy head to compute yaw angles, achieving up to 14 \% higher energy capture than a standard wind-tracking strategy in a low-fidelity steady-state simulator.


\section{Markov Decision Process}\label{sec:markov_decision_process}

A wind farm is a set of \( N \in \mathbb{N}^* \) turbines, indexed by \( i \in \{0, \dots, N-1\} \), each located at fixed spatial coordinates \( (x^i, y^i) \) and characterized by a rotor diameter \( d \) [m].
The control problem is defined over an episode consisting of \( T \in \mathbb{N}^* \) discrete time steps, indexed by \( t \in \{0, \dots, T-1\} \), during which the yaw angle of each turbine is adjusted to optimize performance.
The \gls{wffc} problem is formalized as a \gls{mdp} represented by a tuple \( \langle \mathcal{S}, \mathcal{A}, \mathcal{P}, \mathcal{R}, \gamma \rangle \), with: \( \mathcal{S} \) the set of states, \( \mathcal{A} \) the set of actions, \( \mathcal{P}: \mathcal{S} \times \mathcal{A} \rightarrow \mathcal{S} \) the transition function, \( \mathcal{R}: \mathcal{S} \times \mathcal{A} \rightarrow \mathbb{R} \) the reward function, and \( \gamma \in [0, 1] \) a discount factor.

\subsection{State}\label{ssec:state}

At each time step \( t \), the wind is characterized by a direction \( K_t \in [0, 360] \) degrees and a speed \( V_t \in [V_{\text{min}}, V_{\text{max}}] \) m/s.
This defines the free-stream wind field, which is assumed to be spatially homogeneous across the farm prior to any wake-induced disturbances.
To account for measurement or forecasting uncertainty, we also define \( K_t' \) and \( V_t' \) as the observed or predicted wind direction and speed, respectively, which may differ from the true values due to sensor noise or forecast error.
Each turbine \( i \) has an absolute nacelle orientation \( \beta_t^i \in [0, 360] \) degrees, and a yaw angle \( \alpha_t^i \in [-180, 180] \) degrees.
As described in Sub-Figure~\ref{fig:angles_0}, the yaw angle is the offset between the absolute nacelle orientation and the wind direction, i.e., \( \alpha_t^i = (K_t - \beta_t^i + 180) \mod 360 - 180 \).
The state is \( s_t = (X_{W_t}, X_{F_t}, X_{Y_t}, X_{L_t}) \) where:
\begin{itemize}
	\item \( X_{W_t} = (K_t', V_t') \) represents the current, measured wind conditions;
	\item \( X_{F_t} = ( K_{t+l}', V_{t+l}' )_{l \in \{1, 2, 3\}} \) represents a wind forecast on the next three time steps;
	\item \( X_{Y_t} = (\beta_t^i)_{i \in \{0, \ldots, N-1\}} \) represents the current absolute orientations of each turbine;
	\item \( X_{L_t} = (x^i, y^i, d)_{i \in \{0, \ldots, N-1\}} \) represents the static layout and rotor diameter of each turbine.
\end{itemize}

\subsection{Action}\label{ssec:action}

\begin{figure}[htbp]
	\centering
	\begin{subfigure}[b]{0.31\textwidth}
		\centering
		\includegraphics[width=\textwidth]{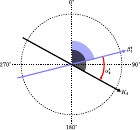}
		\caption{Time step \( t \), start.}\label{fig:angles_0}
	\end{subfigure}
	\hfill
	\begin{subfigure}[b]{0.31\textwidth}
		\centering
		\includegraphics[width=\textwidth]{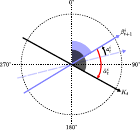}
		\caption{Time step \( t \), update.}\label{fig:angles_1}
	\end{subfigure}
	\hfill
	\begin{subfigure}[b]{0.31\textwidth}
		\centering
		\includegraphics[width=\textwidth]{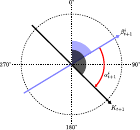}
		\caption{Time step \( t \), end.}\label{fig:angles_2}
	\end{subfigure}
	\caption{Graphical representation of a turbine \( i \) at time step \( t \) with wind direction \( K_t \).
		After action \( a_t^i \) is being applied, the turbine has an updated yaw angle \( \tilde{\alpha}_t^i \).
		At the end of the time step, wind direction evolves from \( K_t \) to \( K_{t+1} \) and the next yaw angle \( \alpha_{t+1}^i \) is computed.}\label{fig:angles}
\end{figure}

The wake steering controller is characterized by a policy \( \pi_\theta \) parameterized by \( \theta \), computing the yaw settings from the state such that \( \pi_\theta(s_t) = \mathbf{a}_t \).
The action \( \mathbf{a}_t = (a_t^i)_{i \in \{0, \ldots, N-1\}} \) is the vector of each turbine individual yaw setting.
Each action \( a_t^i \) corresponds to the rotational movement of turbine \( i \) between time step \( t \) and \( t+1 \) and is bounded in \( [-20, 20] \) degrees due to mechanical constraints of the yaw actuators.
Before action \( a_t^i \) is applied (Figure~\ref{fig:angles_0}), the yaw angle is \( \alpha_t^i \) and the absolute orientation is \( \beta_t^i \).
When action is applied (Figure~\ref{fig:angles_1}), the turbine is rotated, giving an updated orientation \( \beta_{t+1}^i \) and an updated yaw angle \( \tilde{\alpha}_t^i \).
At the end of the time step \( t \) (Figure~\ref{fig:angles_2}), wind direction evolves from \( K_t \) to \( K_{t+1} \) and the next yaw angle \( \alpha_{t+1}^i \) is computed accordingly.

\subsection{Transition}\label{ssec:transition}

At the beginning of each episode, initial wind direction and speed are sampled from uniform distributions such that \( K_0 \sim \mathcal{U}(0, 360) \) and \( V_0 \sim \mathcal{U}(V_{\text{min}}, V_{\text{max}}) \), respectively.
Initial yaw angles are sampled from a uniform distribution such that \( \alpha_0^i \sim \mathcal{U}(-20, 20) \ \forall i \in \{0, \ldots, N-1\} \).
The absolute nacelle orientations \( \beta_0^i \) are then computed based on \( K_0 \).
The wind farm layout remains fixed throughout all episodes.
At each time step \( t \), and for each turbine \( i \):
\begin{enumerate}
	\item the policy computes the yaw settings such that \( \pi_\theta(s_t) = \mathbf{a}_t \);
	\item absolute orientations are updated \( \beta_{t+1}^i = (\beta_t^i + a_t^i) \mod 360 \);
	\item yaw angles are updated \( \tilde{\alpha}_t^i = (K_t - \beta_{t+1}^i + 180) \mod 360 - 180 \);
	\item the power output of the wind farm \( P_t \) in \gls{mw} is computed;
	\item a reward \( r_{t+1} \) is computed, and wind conditions evolve from \( (K_t, V_t) \) to \( (K_{t+1}, V_{t+1}) \);
	\item yaw angles are updated \( \alpha_{t+1}^i = (K_{t+1} - \beta_{t+1}^i + 180) \mod 360 - 180 \);
\end{enumerate}
Power computation is based on the current wind conditions \( (K_t, V_t) \) and the updated yaw angles \( \tilde{\alpha}_t^i \).
In this work, we use a steady-state, low-fidelity wind farm simulator to compute the power outputs and a simple model to generate wind data time series, both later described in Sub-Section~\ref{ssec:experimental_setting}.
To maintain a consistent and valid discretization of the continuous \gls{wffc} problem, we account for turbine rotational constraints by assuming that reorientation occurs rapidly relative to the time step duration, and that wind conditions remain quasi-stationary over each time step.

\subsection{Reward}\label{ssec:reward}

At each time step \( t \), the reward \( r_{t+1} \) (Equation~\ref{eq:r_total}) is the weighted sum of two terms: one for invalid policies and one for power maximization, such that
\begin{gather}
	r_{t+1} = w_0 r_{t+1}^{\text{invalid}} + w_1 r_{t+1}^{\text{power}}. \label{eq:r_total}
\end{gather}

The objective of the \( r_{t+1}^{\text{invalid}} \) term (Equation~\ref{eq:r_invalid}) is to penalize the total reward when some yaws are outside \( [-20, 20] \).
Indeed, outside this interval, turbines are shut down for safety reason, resulting in no power output.
This makes the power maximization reward uninformative.
Therefore, this term is used to guide the policy towards good control strategies, i.e., keeping turbines close to the wind direction.
It is defined as
\begin{gather}
	r_{t+1}^{\text{invalid}} = \frac{-1}{N} \sum_{i=0}^{N-1} \left( \left(\frac{|\tilde{\alpha}^i_t|}{180}\right)^3 \mathbf{1}_{\tilde{\alpha}^i_t \notin [-20, 20]} \right). \label{eq:r_invalid}
\end{gather}

The objective of the \( r_{t+1}^{\text{power}} \) term (Equation~\ref{eq:r_power}) is to maximize power production relative to a baseline, using the power ratio \( \Delta_{P_t} = (P_t - \bar P_t) / \bar P_t \) with \( \bar P_t \) the baseline power output.
Unlike using the absolute power output as a reward - which can bias the agent toward high-production wind conditions only - this ratio normalizes improvements relative to a baseline and ensures the agent optimizes wake steering across all conditions, including those where wake losses are less significant.
We use as a baseline a perfect wind tracking controller that does not perform any wake steering.
It is not subject to the rotational constraints of the turbines, as it always performs simulations with all turbines aligned with the exact wind direction \( K_t \).
The baseline wake losses denoted \( \mathcal{\bar L}_t \) gives some insights about the complexity and importance of \gls{wffc} for the given wind conditions.
High values indicate a significant reduction in energy production, making wake steering essential.

The optimal magnitude of \( \Delta_{P_t} \) depends on the wake losses: a near-zero ratio can be optimal when wake losses are low (making \gls{wffc} unnecessary) but suboptimal when wake losses are high (making \gls{wffc} necessary).
To address this, we introduce an exponential scaling term, parameterized by \( p \), that adjusts the reward based on the baseline wake losses \( \mathcal{\bar L}_t \).
This term ensures a balanced reward across different wind conditions by restricting the power ratio when the wake losses are significant.
Additionally, if power production falls below the baseline, the agent is penalized with a negative reward.
An ablation study is given in Figure~\ref{fig:ablation_study}, in Appendix.
It is defined as
\begin{gather}
	r_{t+1}^{\text{power}} = \Delta_{P_t} \mathbf{1}_{\Delta_{P_t} < 0} + \exp(-p \mathcal{\bar L}_t) \Delta_{P_t} \mathbf{1}_{\Delta_{P_t} \geq 0}. \label{eq:r_power}
\end{gather}


\section{Models}\label{sec:models}

We train and compare three models in a single-agent, continuous action space setting:
a \gls{fnn}-based model named \textbf{V0 model};
a \gls{gat} named \textbf{V1 model};
and our contribution, the \textbf{V2 model}, an attention-based neural network.
Each model follows an actor-critic framework, taking the state \( s_t \) as input and giving both an actor distribution \( (\bar\mu_t, \bar\kappa_t) \) and a critic value \( v \) as outputs.
The actor distribution independently parameterizes a von Mises distribution for each turbine, with the location parameters \( \bar\mu_t := (\mu_t^0, \mu_t^1, \ldots, \mu_t^{N-1}) \) and the concentration parameters \( \bar\kappa_t := (\kappa_t^0, \kappa_t^1, \ldots, \kappa_t^{N-1}) \).

During training, turbine actions are sampled from their respective von Mises distribution: \( a_t^i \sim  \mathcal{V}(\mu_t^i, \kappa_t^i) \ \forall i \in \{0, 1, \ldots, N-1\} \).
During testing, each turbine's action is set directly to its location parameter: \( a_t^i = \mu_t^i, \ \forall i \in \{0, 1, \ldots, N-1\} \).
Due to the inherent symmetry in the \gls{wffc} problem, if an effective solution exists near the lower bound of the action space, a corresponding solution near the upper bound is often equally viable.
By modeling actions with a circular distribution, like the von Mises, we ensure that the policy can explore these equivalent solutions efficiently.
It promotes a more effective and balanced exploration.

To ensure a fair and meaningful comparison, we use approximately the same number of parameters for each model: around 22 million of parameters each.
More details are given in the sub-Section~\ref{ssec:detailed_architectures} of the Appendix.
A hyperbolic tangent activation function, scaled by \( \pi \), is used for \( \bar\mu_t \) to ensure bounded actions in \( [-\pi, \pi] \).
Actions are later denormalized to the turbine's operational range \( [-20, 20] \) before being sent to the environment.
A softplus activation function is used for \( \bar\kappa_t \) to ensure values strictly superior to 1.
A linear activation function is used for the critic output.

\subsection{Model V0}\label{ssec:model_v0}

\begin{figure}[ht]
	\begin{center}
		\includegraphics[width=0.6\textwidth]{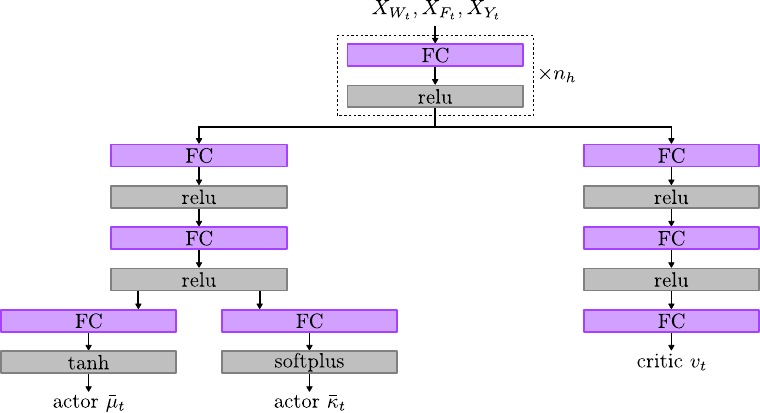}
	\end{center}
	\caption{V0 model, a \gls{fnn}-based architecture.}\label{fig:model_v0}
\end{figure}

The V0 model (Figure~\ref{fig:model_v0}) is a \gls{fnn} composed exclusively of \gls{fc} layers.
Its input is a single concatenated vector comprising the current wind data, wind forecast, and the absolute orientations of the turbines.
The layout feature vector \( X_{L_t} \) is excluded from the inputs as it remains constant across all episodes.
Instead, the model is expected to implicitly learn spatial relationships between turbine coordinates, wind flow, and wake effects during training.
Most existing approaches similarly rely on \glspl{fnn} that process concatenated inputs.
However, this strategy may be inefficient, as it requires the model to simultaneously infer complex spatial and temporal dependencies from a high-dimensional, entangled input vector without explicit structural guidance.

\subsection{Model V1}\label{ssec:model_v1}

\begin{figure}[htbp]
	\centering
	\begin{subfigure}[b]{0.39\textwidth}
		\centering
		\includegraphics[width=\textwidth]{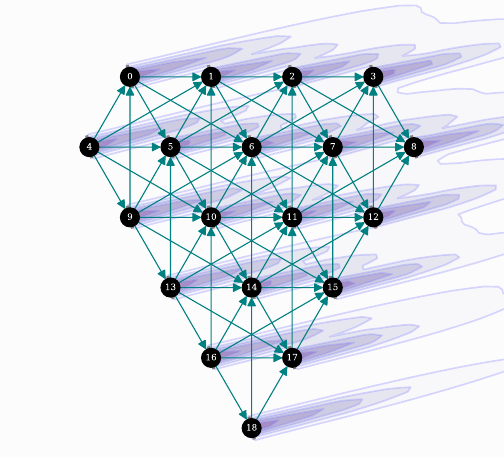}
		\caption{Graph.}\label{fig:graph_figure}
	\end{subfigure}
	\hfill
	\begin{subfigure}[b]{0.59\textwidth}
		\centering
		\includegraphics[width=\textwidth]{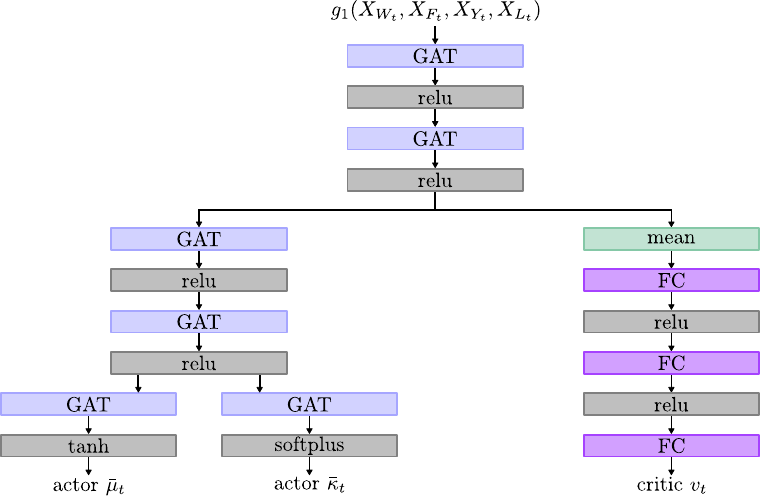}
		\caption{Network.}\label{fig:graph_network}
	\end{subfigure}
	\caption{V1 model, based on \glspl{gat}.
		Sub-Figure~\ref{fig:graph_figure} illustrates an example input graph with a wind direction of 256\textdegree{}, where wake effects are shown in the background.
		Sub-Figure~\ref{fig:graph_network} presents the architecture of the V1 model.}\label{fig:model_v1}
\end{figure}

The V1 model (Figure~\ref{fig:model_v1}) is based on \glspl{gat}~\parencite{graph_attention_networks}.
The input of the V1 model is a graph \( g_1(X_{W_t}, X_{F_t}, X_{Y_t}, X_{L_t}) \) built from the state.
Each node in the graph corresponds to a turbine at a fixed position determined by the layout.
The node feature vector for turbine \( i \) consists in \( (X_{W_t}, X_{F_t}, \beta_t^i) \), which includes current wind data, wind forecast and the turbine's orientation.
A directed edge exists from turbine \( i \) to turbine \( j \) if the distance between them is less than eight turbine diameters and if turbine \( i \) is upstream to \( j \).
We use a distance of eight turbine diameters to balance wake interaction modeling and graph complexity.
Each edge feature vector encodes the normalized distance and relative angle between connected turbines with respect to the wind direction.
\Glspl{gnn} have been successfully applied in deep surrogate modeling~\parencite{learning_to_optimise_wind_farms_with_graph_transformers} and various wind farm analysis tasks but remain relatively underexplored for direct \gls{wffc} optimization.
Still, \glspl{gnn} may not be optimal since certain data (like wind conditions) are duplicated across all turbine nodes, potentially leading to redundancy.

\subsection{Model V2}\label{ssec:model_v2}

\begin{figure}[ht]
	\begin{center}
		\includegraphics[width=0.9\textwidth]{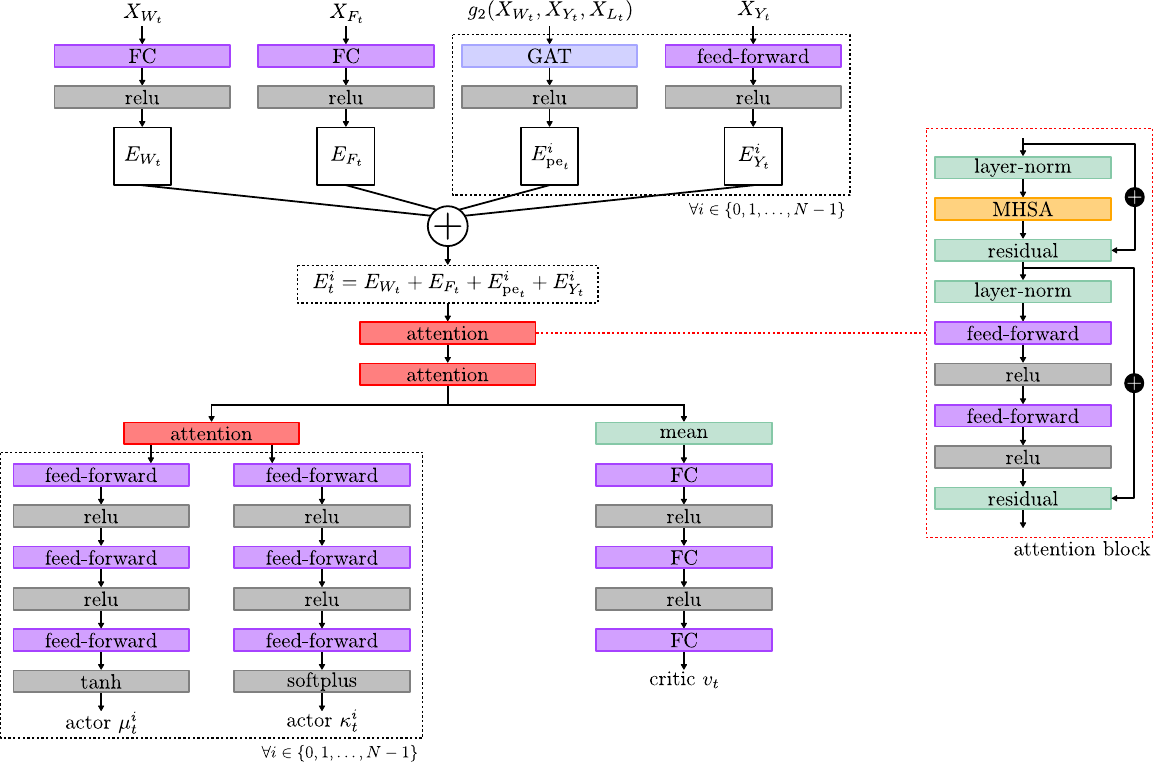}
	\end{center}
	\caption{V2 model architecture, incorporating \gls{gat} and \gls{mhsa} blocks.
		\Gls{fc} layers refer to standard dense layers applied once to the input, while feed-forward layers apply the same \gls{fc} layer independently
		(a) to each turbine's embedding,
		(b) in attention blocks,
		(c) and to each turbine's embedding in the actor branch, after the last attention block.}\label{fig:model_v2}
\end{figure}

Our proposed architecture, described as the V2 model (Figure~\ref{fig:model_v2}), leverages both a \gls{gat} and \gls{mhsa} blocks~\textcite{attention_is_all_you_need}.
To better exploit the multi-modality of the \gls{wffc} problem, inputs are split in four different embeddings.
1) A \gls{fnn} is used to create the wind embedding \( E_{W_t} \).
2) A \gls{fnn} is used to create the forecast embedding \( E_{F_t} \).
3) A \gls{gat} is used to create each turbine positional encoding \( E_{pe_t}^i \).
The input is a graph similar to the one built by the V1 model (Sub-Section~\ref{ssec:model_v1}), without wind speed neither wind forecast.
4) A \gls{fnn} is used to create each turbine specific embedding \( E_{Y_t}^i \) from turbine orientations.
The final embedding of each turbine is the sum of all these embeddings.
Whereas wind and forecast embeddings are shared between all turbines, positional and turbine embeddings are specific for each turbine.
In the context of \gls{wffc}, self-attention captures relationships between turbines by identifying which ones are most relevant for yaw control.
It allows the model to consider all turbines simultaneously and understand how wake effects propagate across the farm.
The multi-head mechanism enhances this by providing multiple perspectives on these interactions.
In natural language processing, positional encoding is straightforward, as it follows word order in a sentence.
In a wind farm, however, turbine positions depend on wind conditions, yaw angles, and wake effects, making encoding more complex.
To address this, we use a \gls{gat} to learn positional embeddings, capturing spatial dependencies more effectively.
By representing the wind farm as a graph, we encode expert knowledge into the model and provide a structured representation of wake interactions, accelerating learning.

\section{Simulations}\label{sec:simulations}

\subsection{Experimental setting}\label{ssec:experimental_setting}

We employ episodic \gls{rl} with a horizon of \( T = 18 \) time steps, each lasting 10 minutes, resulting in a 3 hours control period for the turbines.
The reward loss parameter is set to \( p = 3 \) and the reward weights are \( w_0 =1 \) and \( w_1 = 100 \).
States are normalized to the range \( [-1, 1] \): wind speeds are scaled accordingly and angles are converted to their sine and cosine representations.
We consider a wind farm of \( N = 19 \) turbines and a custom diamond layout as displayed in Figure~\ref{fig:graph_figure}, with a distance of four turbines diameters between a turbine and its closest neighbors.
Numerical simulations are conducted with \gls{floris}~\parencite{floris}, a steady-state, low-fidelity simulator developed by \gls{nrel}.

We consider low speed winds, i.e., \( V_{\text{min}} = 3 \) m/s and \( V_{\text{max}} = 10 \) m/s because this is where \gls{wffc} is the most beneficial for energy production (wake losses have a greater impact).
At each time step step \( t \geq 1 \), we use a simple \gls{arma} process of order 1 to generate wind data.
The direction is computed such that \( K_t = (\epsilon_t + K_{t-1} + 0.1 \epsilon_{t-1}) \mod 360 \) with \( \epsilon_t \sim \mathcal{N}(0, 9) \).
The speed is computed such that \( V_t = \epsilon_t' + V_{t-1} + 0.1 \epsilon_{t-1}' \) with \( \epsilon_t' \sim \mathcal{N}(0, 0.01) \).
A mirroring function is used for the generated speeds to ensure that values stay in \( [V_{\text{min}}, V_{\text{max}}] \).
Noisy wind data is obtained by perturbing the original values with noise sampled from uniform distributions:
\( K_t' = K_t + \epsilon_K \), where \( \epsilon_K \sim \mathcal{U}(-3, 3) \), and \( V_t' = V_t + \epsilon_V \), where \( \epsilon_V \sim \mathcal{U}(-0.1, 0.1) \).

During testing, we use three benchmarks.
\textbf{Standard}: simple benchmark keeping every turbine aligned as possible with the current measured wind direction \( K_t' \).
It does not perform any optimization and does not rely on the wind forecast.
\textbf{\gls{gs}}: good optimized benchmark, introduced by~\textcite{serial_refine_method_for_fast_wake_steering_yaw_optimization}.
It optimizes turbine yaw settings sequentially, where the initial solution is computed with the standard solution.
Then, it sequentially optimizes each turbine from upstream to downstream, keeping the others fixed, by performing a grid search over 40 discretized yaw angles.
It does not rely on the wind forecast.
\textbf{Heuristic}: strong optimized benchmark, introduced by~\textcite{on_the_importance_of_wind_predictions_in_wake_steering_optimization}.
It consists in an improved version of the \gls{gs} solution, where the objective function is augmented by a heuristic.
It does rely on the wind forecast to optimize its solutions on a given horizon, making the comparison with our models more relevant.

\subsection{Proximal policy optimization}\label{ssec:proximal_policy_optimization}

We train each model using a \gls{ppo}~\parencite{proximal_policy_optimization_algorithms} actor-critic method and \gls{gae}~\parencite{high_dimensional_continuous_control_using_generalized_advantage_estimation}.
We use a custom implementation of \gls{ppo} and a custom \gls{rl} environment.
By parallelizing experience collection in our \gls{ppo} implementation and vectorizing power computations in the \gls{floris} simulator, we achieve a 70 speedup in training.
Specifically, training the V2 model takes approximately two hours, whereas without parallelization and vectorization, the same training would require 140 hours.
For reproducibility purpose, more details regarding the training times (Figure~\ref{fig:resources}), the hyperparameters (Table~\ref{tab:hyperparameters_ppo}) and our \gls{ppo} implementation (sub-Section~\ref{ssec:proximal_policy_optimization_appendix}) are given in Appendix.

At each training step, we simulate 360 independent episodes of 3 hours, resulting in 6,480 time steps.
To ensure comprehensive coverage of all possible wind conditions, each of the 360 episodes has a different initial wind direction.
Each episode is initialized with a wind direction sampled from a discrete set of 360 distinct values, uniformly distributed between 0\textdegree{} and 360\textdegree{}.
More specifically, wind directions are sampled in 1\textdegree{} increments, such that the first episode has an initial direction in \( [0, 1] \) degrees, the second episode in \( [1, 2] \), etc.
It ensures that the entire directional space is covered, speeding up generalization and mitigating sampling biases during training.

The discount factor \( \gamma \) determines how much future rewards influence current decisions, where a higher \( \gamma \) prioritizes long-term optimization.
Although our goal is to optimize long-term energy production, we use a small discount factor (\( \gamma = 0.1 \)) due to the steady-state nature of the low-fidelity simulation.
If training were conducted on a higher-fidelity simulation, where state transitions introduce stronger temporal dependencies, increasing the discount factor would be necessary to properly account for long-term effects.

\subsection{Results}\label{ssec:results}

\begin{figure}[ht]
	\begin{center}
		\includegraphics[width=0.65\textwidth]{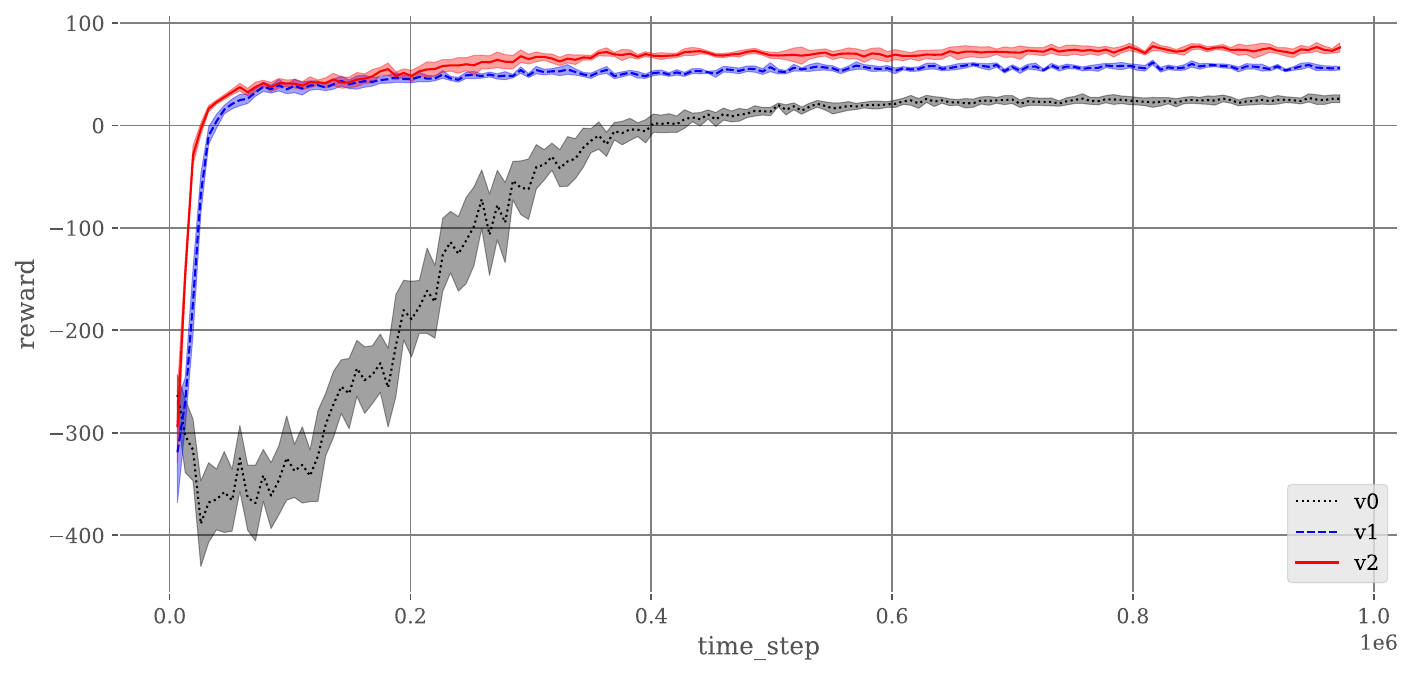}
	\end{center}
	\caption{Training curves of each model, showing mean and variance over 10 different seeds.
		The V1 and V2 models have much faster and stable convergence compared to the V0 model.
		And the V2 model achieves better performance compared to the V0 and V1 models.}\label{fig:train}
\end{figure}

Each model is trained for 150 steps, corresponding to a total of 972,000 simulated time steps.
Training is conducted across 10 different random seeds to ensure robustness and account for variability in learning performance.
The training curves, shown in Figure~\ref{fig:train}, highlight key differences between models.
The V0 model exhibits high variance in the early stages and converges significantly slower than the V1 and V2 models.
Because input data is concatenated into a single vector, the V0 model struggles to learn an effective policy efficiently, likely due to the lack of an explicit structural representation of the data, requiring more training steps to stabilize.
In contrast, the V2 model demonstrates superior stability and faster convergence, ultimately outperforming both the V0 and V1 models in terms of learning efficiency and final performance.

\begin{figure}[htbp]
	\centering
	\begin{subfigure}[b]{0.31\textwidth}
		\centering
		\includegraphics[width=\textwidth]{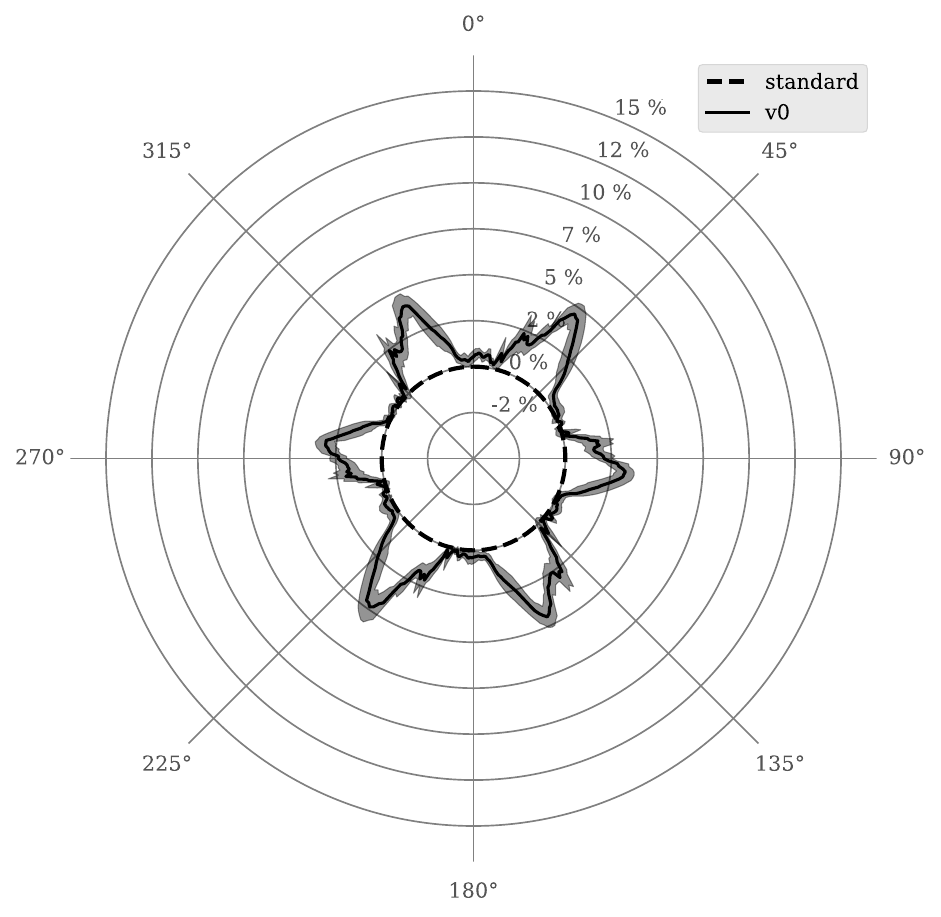}
		\caption{V0 model.}\label{fig:test_v0}
	\end{subfigure}
	\hfill
	\begin{subfigure}[b]{0.31\textwidth}
		\centering
		\includegraphics[width=\textwidth]{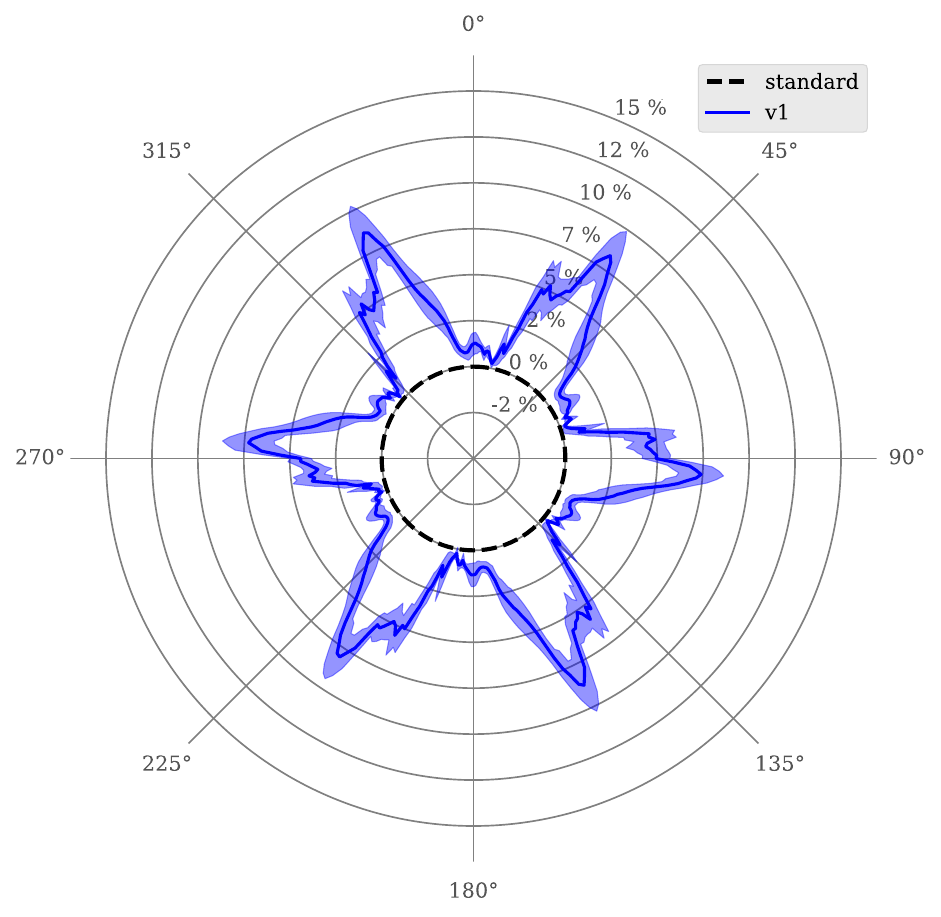}
		\caption{V1 model.}\label{fig:test_v1}
	\end{subfigure}
	\hfill
	\begin{subfigure}[b]{0.31\textwidth}
		\centering
		\includegraphics[width=\textwidth]{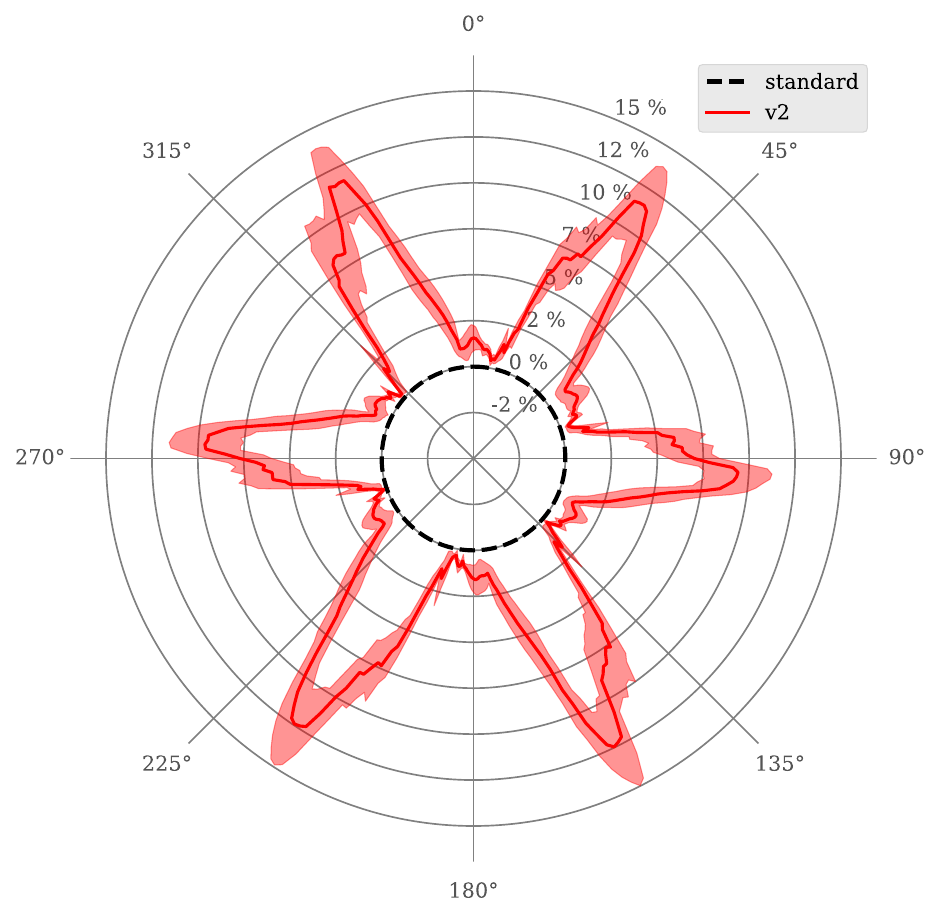}
		\caption{V2 model.}\label{fig:test_v2}
	\end{subfigure}
	\hfill
	\vspace{-0pt}
	\begin{subfigure}[b]{0.31\textwidth}
		\centering
		\includegraphics[width=\textwidth]{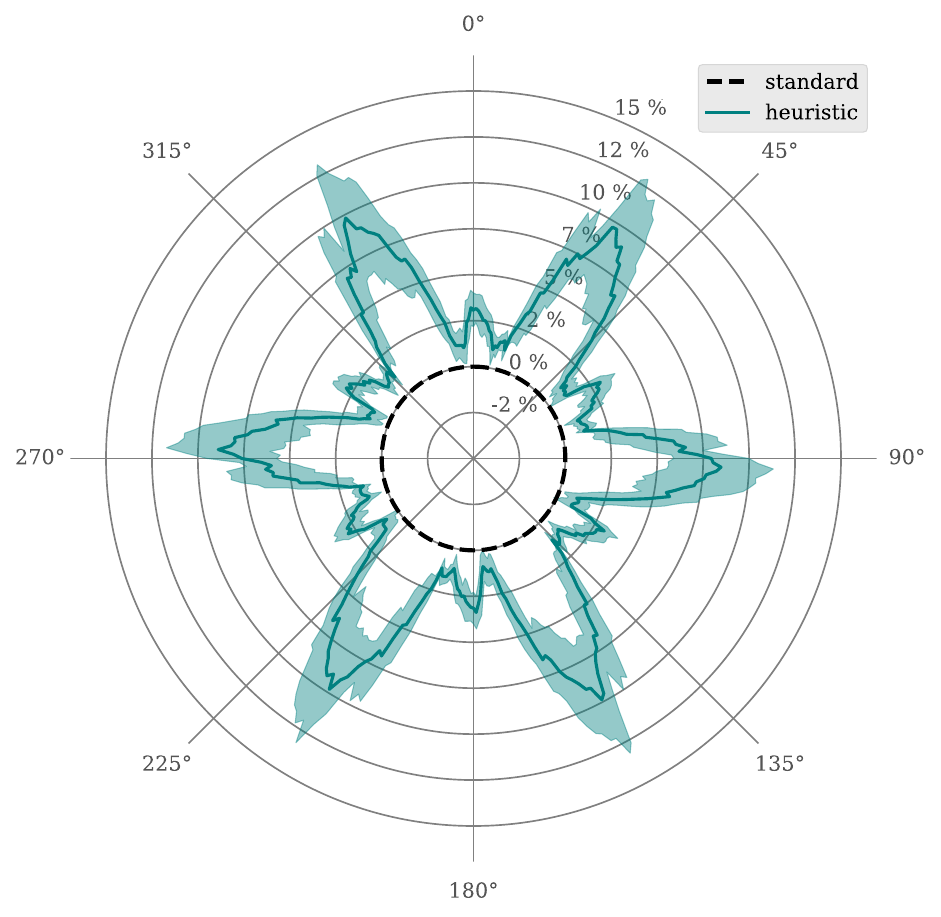}
		\caption{Heuristic.}\label{fig:test_heuristic}
	\end{subfigure}
	\hspace{11pt}
	\begin{subfigure}[b]{0.31\textwidth}
		\centering
		\includegraphics[width=\textwidth]{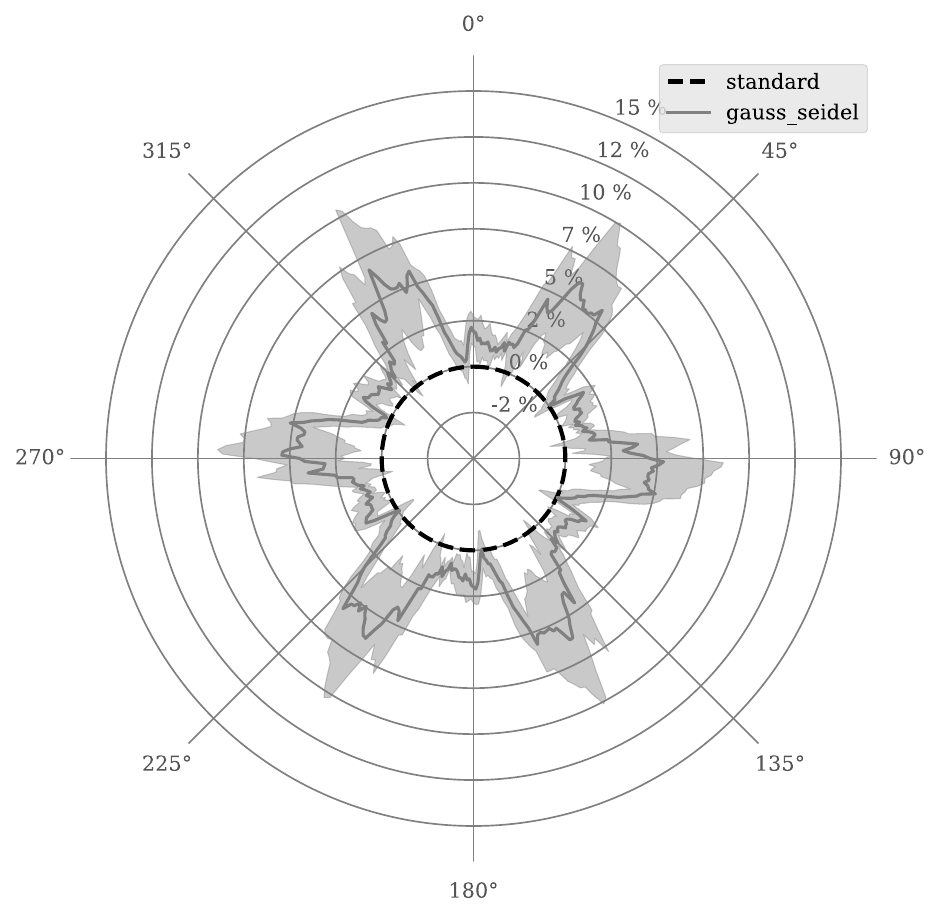}
		\caption{Gauss-Seidel.}\label{fig:test_gauss_seidel}
	\end{subfigure}
	\caption{Performance of each solution relative to the standard benchmark.
		The V2 model significantly outperforms the V0 and V1 models as well as the \gls{gs} solution.
		It achieves performance comparable to the heuristic, but with lower variance and higher gains in strong wake loss conditions.}\label{fig:test}
\end{figure}

To evaluate the generalization of each solution across all wind conditions, we test them on 360 wind directions, sampled in 1-degree increments.
For each direction, we run 10 independent test episodes of 18 time steps, using random seeds not used during training.
This ensures that test episodes remain distinct from training and provide comprehensive directional coverage.
For each episode, we compute each solution's cumulative power production and quantify its improvement over the standard wind-tracking solution.
We then report the mean and variance of these improvements across the 10 seeds for each wind direction and present the results in Figure~\ref{fig:test}.

The V2 model consistently increases wind farm energy production, achieving gains of up to 14 \% in high wake-loss scenarios.
It outperforms both the V0 and V1 models across nearly all conditions.
Although the heuristic baseline delivers strong performance, its results exhibit greater variance, making it less reliable.
The \gls{gs} approach performs poorly and, in some cases, even yields lower power output than the standard solution.
This demonstrates that wind direction can shift too rapidly relative to yaw constraints, making long-term optimization essential for effective control.
The V2 model successfully leverages noisy wind forecasts to improve long-term performance.
Notably, in strong wake conditions, the V2 model outperforms the heuristic controller while being roughly 200 times more computationally efficient.
However, when wake losses are low, power gains are more limited, and the V2 model underperforms compared to the heuristic.
The underlying cause of this discrepancy remains unclear: it may come from architectural limitations or from the reward function design.
Future works should investigate these factors to further refine the model's performance.

\section{Conclusion}\label{sec:conclusion}

In this work, we introduced a novel deep \gls{rl} architecture for wake steering, based on \glspl{gat} and \gls{mhsa}, improving by approximately a factor 10 the sampling efficiency of a \gls{fnn}.
Once trained with \gls{ppo}, our model computes the yaw angles of each turbine and achieves more robust performance than a strong optimization baseline, increasing energy production by up to 14 \%.
To the best of our knowledge, this work is the first to achieve complete generalization over time-varying wind conditions, thanks to a novel reward function and training strategy.
However, while this work provides encouraging evidence of the potential for deep \gls{rl} for robust wake steering, the results remain empirical and are based on simplified, steady-state, low-fidelity wake models.
For real-world deployment, future work should incorporate turbine fatigue considerations to ensure long-term structural integrity and validate the approach in higher-fidelity and unsteady flow environments that better capture realistic wind dynamics.
As the problem grows in complexity, the strengths of \gls{rl} - such as its ability to optimize over long horizons, adapt to uncertain dynamics, and operate without explicit system models - should further reinforce its suitability for wake steering control.

\printbibliography


\newpage
\appendix
\section{Appendix}

\subsection{Detailed architectures}\label{ssec:detailed_architectures}

\subsubsection*{V0 model}

The V0 model, a \gls{fnn}-based architecture, is presented in Figure~\ref{fig:model_v0}.
In this work, \( n_h = 2 \) and the two shared \gls{fc} layers have output sizes of 1024 and 4096.
The shared actor branch comprises two \gls{fc} layers with output sizes of 2048 and 256.
The \( \bar\mu_t \) and \( \bar\kappa_t \) actor branches each contain a single \gls{fc} layer with output size of \( N \).
The critic branch consists of three \gls{fc} layers with output sizes of 2048, 256, and 1.

\subsubsection*{V1 model}

The V1 model, a \gls{gat}, is presented in Figure~\ref{fig:model_v1}.
In this work, the two shared \gls{gat} layers have three attention heads and output sizes of 1024.
The shared actor branch comprises two \gls{gat} layers with three attention heads and output sizes of 128 and 64.
The \( \bar\mu_t \) and \( \bar\kappa_t \) actor branches have both a single \gls{gat} layer with one attention head, giving a scalar output for each node (i.e., for each turbine).
The critic branch consists of three \gls{fc} layers with output of sizes 128, 64, and 1.

\subsubsection*{V2 model}

The V2 model, an attention-based neural network, is presented in Figure~\ref{fig:model_v2}.
In this work, each embedding layer has an output size of 256 and the \gls{gat} used for positional encoding has a single attention head.
The three attention blocks consist of a \gls{mhsa} layer with three attention heads and an output size of 256, followed by two feed-forward layers: one increases the size four times and the other restores it.
Both actor branches contain three feed-forward layers with output sizes of 128, 64, and 1.
The critic branch follows the same structure, with three \gls{fc} layers of output sizes 128, 64, and~1.

\subsection{Training times}\label{ssec:training_times}

\begin{figure}[ht]
	\begin{center}
		\includegraphics[width=0.8\textwidth]{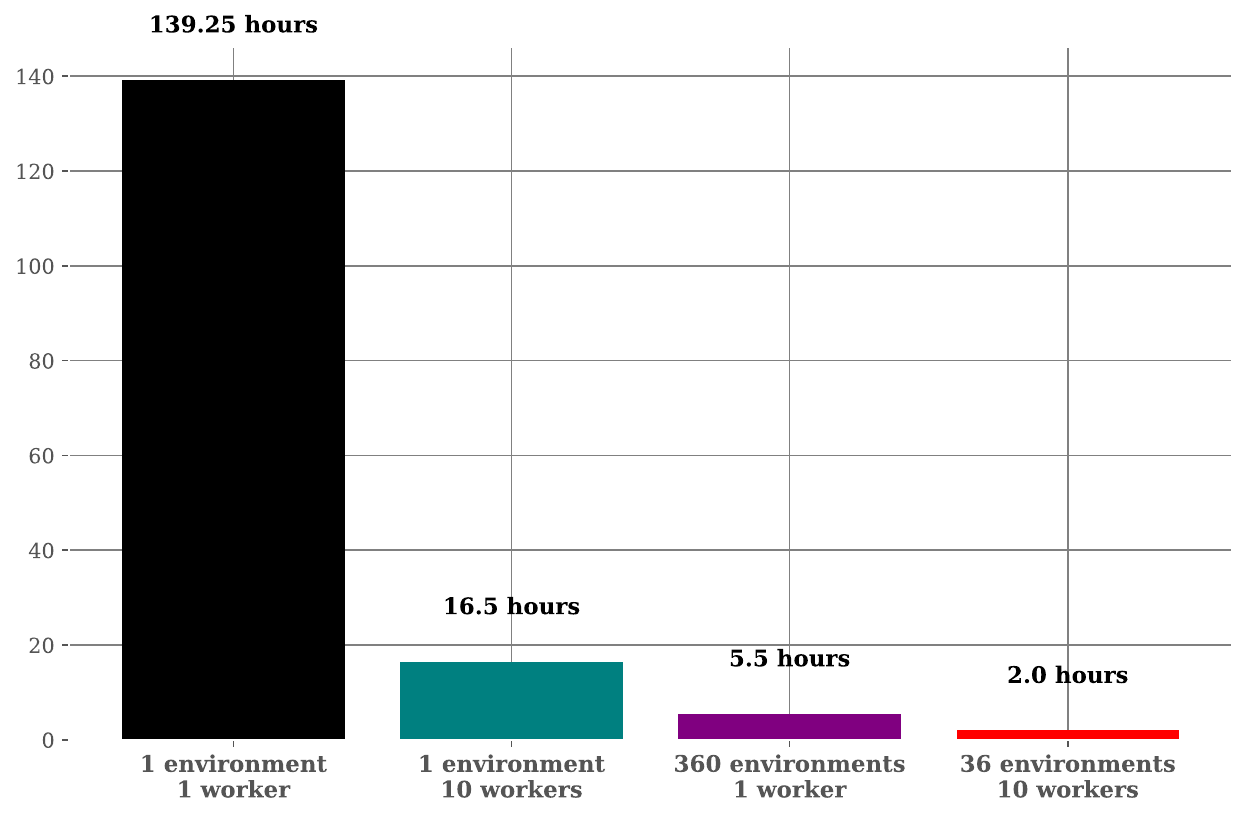}
	\end{center}
	\caption{Training time of the V2 model under different resource configurations.
		Parallelizing experience collection in \gls{ppo} and vectorizing power computations in the \gls{floris} simulator leads to a 70 speedup for training.
		Each model is trained on a NVIDIA Grace \gls{cpu} and NVIDIA GH200 Hopper \gls{gpu}.}\label{fig:resources}
\end{figure}

\subsection{Hyperparameters}\label{ssec:hyperparameters}

Numerical simulations are conducted with \gls{floris}~\parencite{floris}, a steady-state, low-fidelity simulator developed by \gls{nrel}.
The default Gaussian-curl hybrid model~\parencite{control_oriented_model_for_secondary_effects_of_wake_steering} provided by \gls{floris} is used.
The machines are \gls{iea} 15 \glspl{mw} wind turbines~\parencite{definition_of_the_iea_wind_15_megawatt_offshore_reference_wind_turbine}.
For the \gls{ppo}, hyperparameters are listed in Table~\ref{tab:hyperparameters_ppo}.

\begin{table}[htbp]
	\caption{\Gls{ppo} hyperparameters used for each model.
		Only the learning rates and gradient clipping parameters differ.
		At each training step, the current learning rate is sampled using linear interpolation between the initial and final values.}\label{tab:hyperparameters_ppo}
	\centering
	\begin{tabular}{llll}
		\\
		\toprule
		\multicolumn{1}{l}{\bf Hyperparameter} & \multicolumn{1}{l}{\bf Model V0} & \multicolumn{1}{l}{\bf Model V1} & \multicolumn{1}{l}{\bf Model V2} \\

		\cmidrule(r){1-4}
		Training steps                         & 150                              & 150                              & 150                              \\
		Discount factor                        & 0.1                              & 0.1                              & 0.1                              \\
		Learning rate (first)                  & 1e-4                             & 1e-5                             & 1e-5                             \\
		Learning rate (last)                   & 1e-6                             & 1e-7                             & 1e-7                             \\
		Gradient clipping                      & 10                               & None                             & None                             \\
		\Gls{gae} \( \lambda \) parameter      & 0.95                             & 0.95                             & 0.95                             \\
		Entropy coefficient                    & 0.05                             & 0.05                             & 0.05                             \\
		Clip parameter (actor)                 & 0.01                             & 0.01                             & 0.01                             \\
		Clip parameter (critic)                & 10                               & 10                               & 10                               \\
		Value loss coefficient                 & 0.1                              & 0.1                              & 0.1                              \\
		Number of epochs                       & 11                               & 11                               & 11                               \\
		Train batch size                       & 6480                             & 6480                             & 6480                             \\
		Mini batch size                        & 360                              & 360                              & 360                              \\
		\bottomrule
	\end{tabular}
\end{table}

\subsection{Ablation study}

\begin{figure}[htbp]
	\centering
	\begin{subfigure}[b]{0.32\textwidth}
		\centering
		\includegraphics[width=\textwidth]{./resources/test_v2.pdf}
		\caption{Default.}\label{fig:test_v2_default}
	\end{subfigure}
	\hfill
	\begin{subfigure}[b]{0.32\textwidth}
		\centering
		\includegraphics[width=\textwidth]{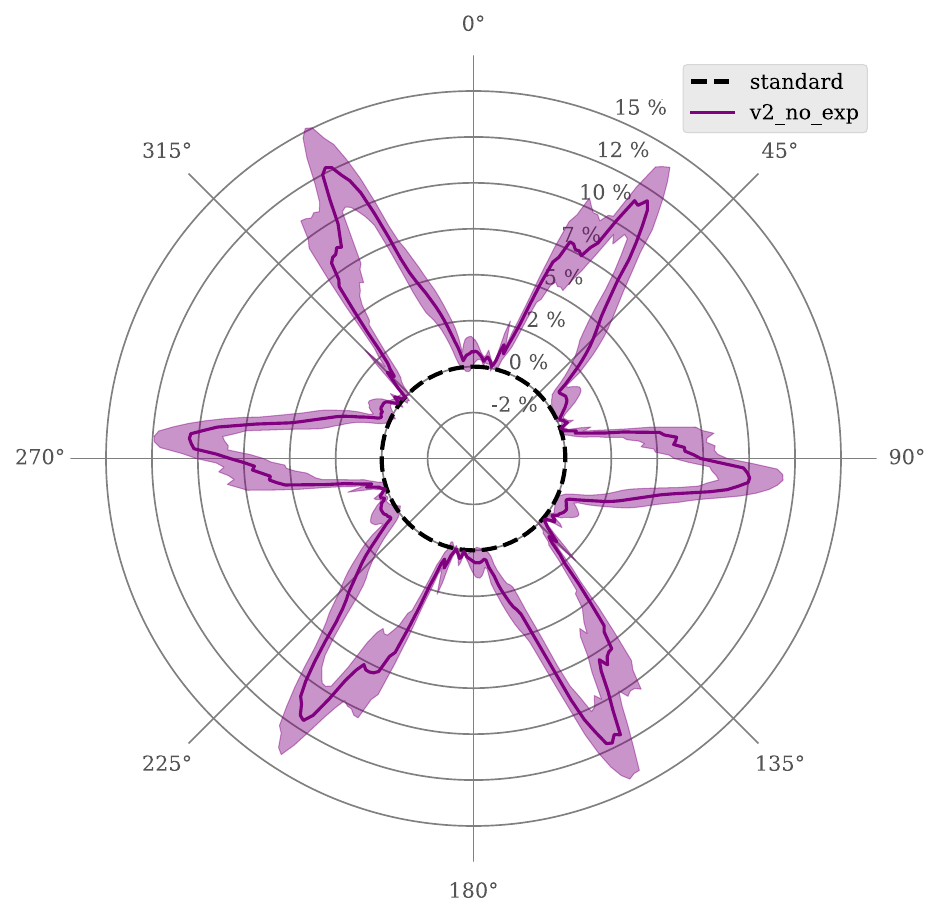}
		\caption{Without \( \exp(-p \mathcal{\bar L}_t) \).}\label{fig:test_v2_no_exp}
	\end{subfigure}
	\hfill
	\begin{subfigure}[b]{0.32\textwidth}
		\centering
		\includegraphics[width=\textwidth]{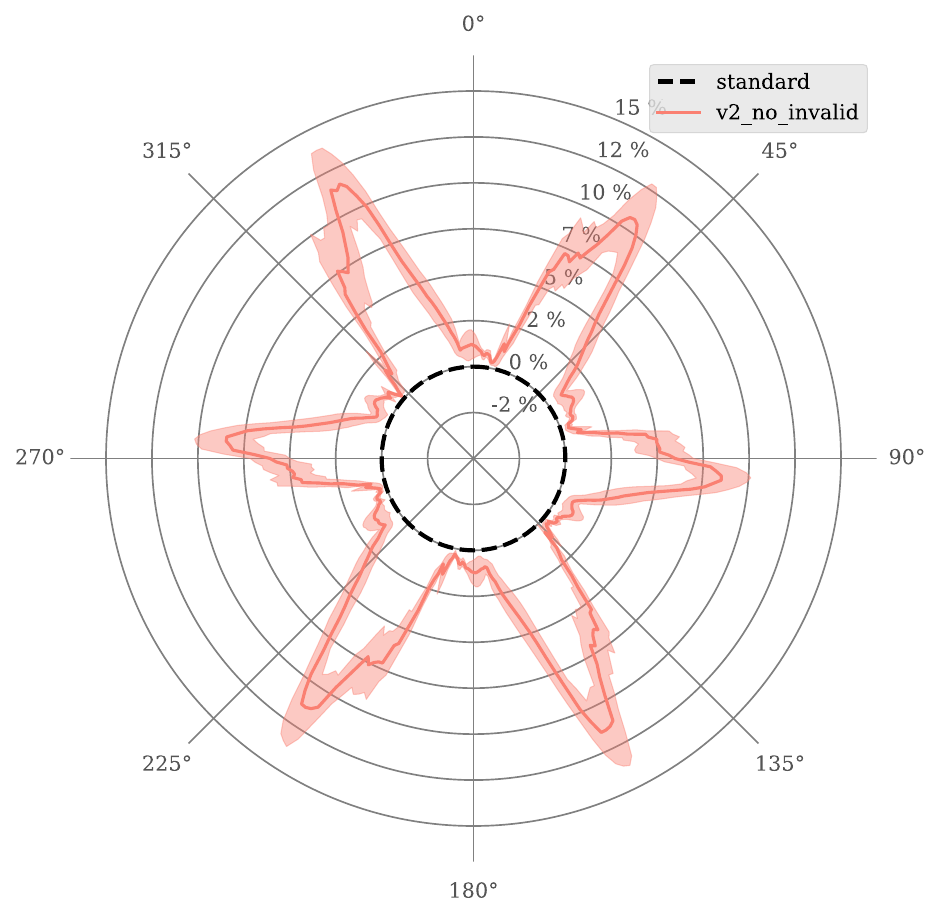}
		\caption{Without \( r_{t+1}^{\text{invalid}} \).}\label{fig:test_v2_no_invalid}
	\end{subfigure}
	\caption{Ablation study for the V2 model.
		Without the exponential term, i.e., with \( p = 0 \), performance is degraded for small wake losses conditions.
		And without the invalid term, i.e., with \( w_0 = 0 \), performance is globally decreased.}\label{fig:ablation_study}
\end{figure}

We conduct an ablation study to assess the influence of the exponential and invalid terms within our reward function.
Figure~\ref{fig:test_v2_default} displays the performance of the default V2 model.
In Figure~\ref{fig:test_v2_no_exp}, we evaluate the V2 model trained with \( p = 0 \), effectively removing the exponential term's impact.
And Figure~\ref{fig:test_v2_no_invalid} shows results when the \( r_{t+1}^{\text{invalid}} \) term is eliminated by setting \( w_0 = 0 \).
The study indicates that the exponential term is crucial for consistent optimization across all wind conditions, as its absence leads to suboptimal performance in scenarios with small wake losses.
Furthermore, the invalid term contributes to overall performance, with its removal resulting in a slight global degradation.

\subsection{Proximal policy optimization}\label{ssec:proximal_policy_optimization_appendix}

\Gls{ppo}~\parencite{proximal_policy_optimization_algorithms} is a model-free, on-policy deep \gls{rl} algorithm based on policy gradients.
It improves stability and sample efficiency over standard policy gradient methods by introducing a clipped surrogate objective that constrains policy updates.
In this work, we have developed a custom actor-critic implementation of \gls{ppo}, drawing inspiration from Stable Baselines 3~\parencite{stable_baselines3} and RLlib~\parencite{rllib_2018,rllib_2021}.
Our implementation is fully vectorized using NumPy and PyTorch matrix operations~\parencite{array_programming_with_numpy}, and the collection of trajectories is fully parallelized via the Ray library~\parencite{ray} to enable efficient large-scale training.
It supports both continuous and discrete action spaces.
Let \( \pi_{\theta} \) and \( V_{\theta} \) be a policy and value function, respectively, with shared parameters \( \theta \).
The pseudocode of our \gls{ppo} training loop is given in Algorithm~\ref{alg:pseudocode_of_ppo_training_loop}.
It consists of two distinct phases: the collection of trajectories and the update of the parameters.

During training, trajectories are collected by executing the current policy \( \pi_\theta \) and the current value \( V_\theta \) in the environment.
These trajectories consist of tuples \( (s_t, a_t, r_{t+1}, s_{t+1}) \) over multiple time steps.
At the end of a trajectory, the value function is used to bootstrap the final return only when the episode is truncated (e.g., due to time limits).
In that case, the final estimated value \( V_\theta(s_T) \) is used as a proxy for future rewards: \( r_T \leftarrow r_T + \gamma V_\theta(s_T) \).
This ensures consistent return estimation across both terminated and truncated episodes.
At the end of an episode, \gls{gae}~\parencite{high_dimensional_continuous_control_using_generalized_advantage_estimation} is used to compute the value targets \( \hat{V}_t \) and the advantages \( \hat{A}_t \).
This method provides a biased but low-variance estimator of the advantages.
It relied on the \gls{td} residual \( \delta_t = r_t + \gamma V_\theta(s_{t+1}) - V_\theta(s_t) \).
The advantages are computed as an exponentially-weighted sum \( \hat{A}_t = \sum_{k=0}^{T-t-1} (\gamma \lambda)^k \delta_{t+k} \) with \( \lambda \) the \gls{gae} parameter.
And the value targets are estimated as \( \hat{V}_t = \hat{A}_t + V_\theta(s_t) \).

After collecting multiple trajectories, the training process enters an optimization phase consisting of several epochs.
During each epoch, the full batch of collected data is shuffled and divided into minibatches.
Each minibatch is then used to compute the loss and update the parameters of both the policy and value networks via \gls{sgd}.
The total loss is defined as a weighted sum of four components:
\( \mathcal{L} = c_0 \mathcal{L}_{\text{actor}} + c_1 \mathcal{L}_{\text{critic}} + c_2 \mathcal{L}_{\text{entropy}}, \)
where \( c_0, c_1 \), and \( c_2 \) are scalar hyperparameters controlling the relative contribution of each term.
The probability ratio between the current and old policy is given by \( r_t(\theta) = \pi_{\theta}(a_t|s_t) / \pi_{\theta_{\text{old}}}(a_t|s_t) \) and is central to the actor loss.
The four loss terms are defined as follows.
\begin{itemize}
	\item \textbf{Actor loss:}
	      \( \mathcal{L}_{\text{actor}} = - \min\left( r_t(\theta)\hat{A}_t, \text{clip}(r_t(\theta), 1 - \varepsilon, 1 + \varepsilon)\hat{A}_t \right) \),
	      where \( \hat{A}_t \) is the advantage estimate and \( \varepsilon \in (0,1) \) is a trust-region hyperparameter.
	      The clipping prevents large policy updates that could destabilize training.
	\item \textbf{Critic loss:}
	      \( \mathcal{L}_{\text{critic}} = \text{clip}\left( (V_\theta(s_t) - \hat{V}_t)^2, 0, \texttt{vf\_clip\_param} \right) \),
	      where \( \hat{V}_t \) is the target return and \texttt{vf\_clip\_param} controls the maximum contribution of the value error to the total loss, ensuring stability in value updates.
	\item \textbf{Entropy loss:}
	      \( \mathcal{L}_{\text{entropy}} = - \mathcal{H}(\pi_\theta(s_t)) \),
	      where \( \mathcal{H} \) denotes the Shannon entropy of the policy.
	      This term encourages exploration by penalizing low-entropy (overly deterministic) policies.
\end{itemize}

\begin{algorithm}[H]
	\footnotesize
	\floatname{algorithm}{Algorithm}
	\algrenewcommand\algorithmicrequire{\textbf{Input: }}
	\algrenewcommand\algorithmicensure{\textbf{Output: }}
	\caption{Pseudocode of \acrshort{ppo} training loop.}\label{alg:pseudocode_of_ppo_training_loop}
	\begin{algorithmic}[1]
		\Require Initial policy \( \pi_\theta \) and value \( V_\theta \)
		\Require \Gls{ppo} hyper-parameters
		\For{\texttt{iter} = 0 to \texttt{max\_iters}}
		\State Collect trajectories using \( \pi_{\theta} \)
		\State Compute value targets \( \hat{V}_t \)
		\State Compute advantages \( \hat{A}_t \)
		\State Fix parameters \( \theta_{\text{old}} \longleftarrow \theta \)
		\For{\texttt{epoch} = 0 to \texttt{nb\_epochs}}
		\For{\texttt{minibatch} in \texttt{batches}}
		\State Compute ratio \( r_t = \pi_\theta(a_t|s_t)/\pi_{\theta_{\text{old}}}(a_t|s_t) \)
		\State Compute loss \( \mathcal{L} = c_0 \mathcal{L}_{\text{actor}} + c_1 \mathcal{L}_{\text{critic}} + c_2 \mathcal{L}_{\text{entropy}} \)
		\State Update parameters \( \theta \longleftarrow \theta - \alpha \nabla_\theta \mathcal{L} \)
		\EndFor
		\EndFor
		\EndFor
	\end{algorithmic}
\end{algorithm}

The policy outputs \( p_0 = (p_0^i)_{i \in \{0, \ldots, N-1\}} \) and \( p_1 = (p_1^i)_{i \in \{0, \ldots, N-1\}} \), which parameterize a set of independent von Mises distributions, one per turbine.
For each turbine \( i \), the location parameter is defined as \( \mu^i = \pi \cdot \tanh(p_0^i) \) and the concentration parameter as \( \kappa^i = \text{softplus}(p_1^i) \).
During training, actions are sampled independently for each turbine as \( a_t^i \sim \mathcal{V}(\mu^i, \kappa^i) \), while during evaluation, actions are set deterministically to the mode, \( a_t^i = \mu^i \).
The use of von Mises distributions ensures that the yaw angles are naturally constrained within the circular interval \( [-\pi, \pi] \).

As the von Mises distribution is rarely used as a policy output in continuous \gls{rl}, we provide additional details below.
Specifically, we present the expressions for its probability density function and entropy.
The probability density function is defined as \( f(x \mid \mu, \kappa)=\frac{\exp(\kappa \cos(x-\mu))}{2 \pi I_0(\kappa)} \),
with \( \mu \) the measure of location, \( \kappa \) the measure of concentration and \( I_0(k) \) the modified Bessel function of the first kind of order 0.
The Bessel's integral \( I_n(\kappa) \) can be written as \( I_n(\kappa) = \frac{1}{\pi} \int_0^\pi \cos(n x) \exp(\kappa \cos(x)) \ dx \).
The entropy \( \mathcal{H}(f) \) is computed as follows.
\begin{align}
	\mathcal{H}(f) & =-\int_{-\pi}^\pi f(x) \log(f(x)) \ dx                                                                                                                \\
	               & =-\int_{-\pi}^\pi f(x)(\kappa \cos(x - \mu) - \log(2 \pi I_0(\kappa))) \ dx                                                                           \\
	               & =-\int_{-\pi}^\pi f(x) \kappa \cos(x - \mu) \ dx + \int_{-\pi}^\pi f(x) \log(2 \pi I_0(\kappa)) \ dx                                                  \\
	               & = \frac{-\kappa}{2 \pi I_0(\kappa)} \int_{-\pi}^\pi \cos(x - \mu) \exp(\kappa \cos(x - \mu)) \ dx + \log(2 \pi I_0(\kappa)) \int_{-\pi}^\pi f(x) \ dx \\
	               & = \frac{-\kappa}{2 \pi I_0(\kappa)} 2 \pi I_1(\kappa)+\log \left(2 \pi I_0(\kappa)\right)                                                             \\
	               & = -\kappa \frac{I_1(\kappa)}{I_0(\kappa)} + \log \left(2 \pi I_0(\kappa)\right).
\end{align}

\end{document}